\documentclass[letterpaper, 10 pt, conference]{ieeeconf}

\IEEEoverridecommandlockouts
\usepackage{graphicx}
\usepackage{amsmath}
\usepackage{booktabs}
\usepackage{xcolor}
\ifdefined\anonymoussubmission \else \def\publishversion{} \fi
\ifdefined\publishversion \def\hiderevmarks{} \fi
\ifdefined\hiderevmarks
  \newcommand{\rev}[1]{#1}
\else
  \newcommand{\rev}[1]{#1}
\fi
\ifdefined\hidenamedmarks
  \newcommand{\named}[1]{#1}
\else
  \newcommand{\named}[1]{#1}
\fi
\usepackage{cite}
\makeatletter
\renewcommand{\citepunct}{],\penalty\@m\ [}

\makeatother

\graphicspath{{figures/}{../figures/}}   

\title{\LARGE \bf
Swim-and-Breach at Palm Scale: A Rudder-Steered Two-Propeller
Underwater Robot Platform with Differential-Thrust Pitch Control}

\newif\ifanon
\ifdefined\publishversion \anonfalse \else \anontrue \fi

\ifanon
\author{Anonymous submission for double-anonymous review}
\else
\author{\named{Daehyun Choi$^{1,*,\dagger}$, Ian Bergerson$^{1,*}$, Hengjia Zhu$^{1,*}$,
Tianjun Lan$^{1}$, and Saad Bhamla$^{1,\dagger}$}%
\thanks{\named{$^{*}$These authors contributed equally to this work.}}%
\thanks{\named{$^{\dagger}$Corresponding authors.}}%
\thanks{\named{$^{1}$The authors are with the BioFrontiers Institute, University of
        Colorado Boulder, Boulder, CO 80309, USA.
        {\{daehyun.choi, saad.bhamla\}@colorado.edu}}}%
}
\fi

\begin{document}

\maketitle
\thispagestyle{empty}
\pagestyle{empty}

\begin{abstract}

We present a palm-scale (65~mm, 34~g) swim-and-breach robot platform.
Two vertically stacked propellers provide both propulsion and
differential-thrust pitch control under a proportional-integral-derivative (PID) loop, and a tail
rudder adds yaw control. The hull, evaluated by flow simulation, reduces
the drag five-fold relative to an equivalent cuboid, and the propellers
are optimized using B-series modeling validated by
dynamometer measurements. The current robot swims at 13.9
body lengths per second and turns at 209 deg per second, corresponding to the upper limits reported for underwater robots. In free swimming, the pitch loop turns the body to
any commanded nose-up pitch angle, and, with the rudder
stabilizing the exit, the current robot leaps 1.6 body lengths high and 3.7 long in a
seamless cruise-leap-cruise sequence.
The platform can be used to build small-scale
robots that cross barriers and dry gaps between pools for inspection
in streams, flooded structures, and industrial systems.
\end{abstract}

\section{INTRODUCTION}

\begin{figure}[t]
  \centering
  \includegraphics[width=0.95\columnwidth]{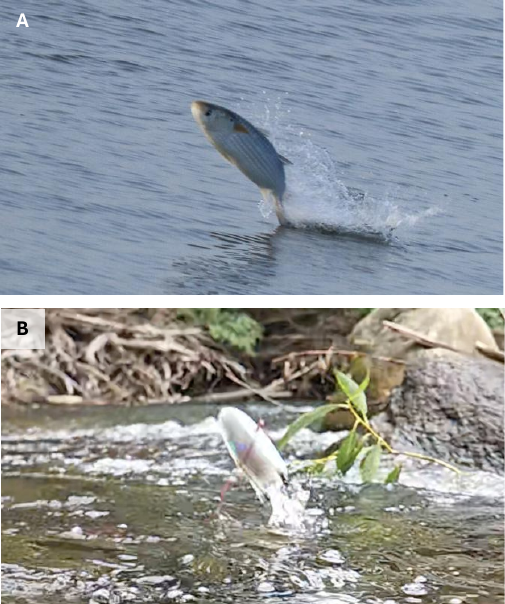}
  \caption{Water exit of (A)~a flathead grey mullet (\emph{Mugil cephalus};
  photo by Afsarnayakkan, Wikimedia Commons, CC~BY-SA~4.0) and (B)~the
  current robot in \ifanon a natural stream\else \named{Boulder Creek, Colorado}\fi{}.}
  \label{fig:intro}
\end{figure}

To leave the water from a level swim, a fish pitches nose-up,
accelerates toward the surface with a burst of thrust, and holds that
pitch until it breaks through the free surface
(Fig.~\ref{fig:intro}(A)). Fishes across
taxa leap for multiple purposes: outrunning predators
below~\cite{davenport1994flyingfish}, clearing waterfalls in
migration~\cite{lauritzen2010salmon}, shedding
parasites~\cite{dempster2019sealice}, capturing prey above the
surface~\cite{shih2017archerfish}, and escaping a sudden
disturbance~\cite{vetter2017silvercarp}.

Underwater robots have matured as swimmers~\cite{wang2022survey}. Soft
bodies cruise with fishlike motion~\cite{katzschmann2018sofi},
high-frequency tail actuation approaches biological swimming
performance~\cite{zhu2019tunabot}, magnetic actuation has produced
maneuverable designs~\cite{chen2020magnetic}, and
wire-driven mechanisms offer direction control in compact
packages~\cite{wang2026wiredriven}. The agility of these robots,
however, is reported
almost entirely as turning in the horizontal plane (perpendicular to gravity), whereas a water
exit takes place in the vertical plane (parallel to gravity) and \rev{requires} a hydrodynamic
design built for burst thrust as well as fast and precise pitch and yaw
control at the instant of exit.

\begin{figure}[t]
  \centering
  \includegraphics[width=0.95\columnwidth]{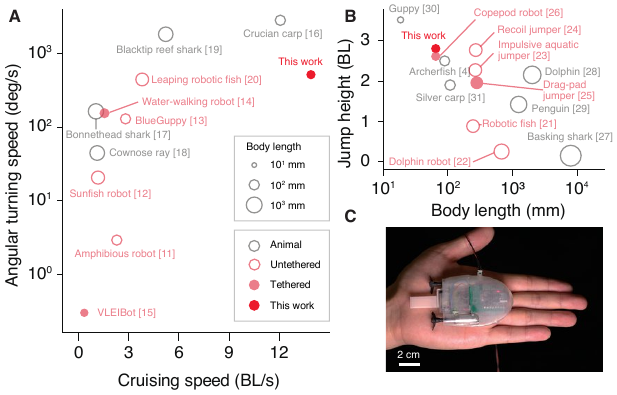}
  \caption{\rev{Performance of the current robot against published robots and animals. (A)~Reported angular turning rate (deg/s) against forward swimming speed normalized by body length (BL/s). (B)~Jump height above the water surface against body length. In both panels, open and filled light red circles denote untethered and tethered or mechanically supported robots, open black circles denote animals, and the filled red circle denotes the current robot. Each label gives the name and the reference number, and the circle diameter scales with the logarithm of the body length. (C)~The current robot on a palm.}}
  \label{fig:compare}
\end{figure}

\rev{To replicate the fish-inspired swim-and-breach maneuver, we developed a robot platform
(Fig.~\ref{fig:intro}(B)).}
\rev{We chose propellers as thrusters over a flapping tail because propellers are known to
deliver at least an order of magnitude greater static thrust per unit
weight~\cite{trygstad2025force,hartmann2025flat}.}
\rev{We vertically stacked the propellers for two reasons: (i) simple pitch control through the
thrust difference between the two propellers, with a rudder mechanism correcting yaw deviations; and (ii) a low-drag design by positioning the motors along the same vertical
center plane so that the body can be kept laterally thin.}
\rev{Both in the laboratory and in a natural stream, we showed, through video tracking, that the current platform
swam at 13.9 body lengths per second (BL/s), placing it among the fastest existing robotic
systems~\cite{wang2022amphibious,huang2025sunfish,ko2025blueguppy,kim2022waterwalking,blankenship2024vleibot,chen2022tmech}
in terms of both swimming speed and turning rate, and among the fastest even of the
fish compared here~\cite{domenici2008crucian,hoffmann2019bonnethead,parson2011batoids,trujillo2022sharks},
while its pitching rate is comparable to that of the sharks
(Fig.~\ref{fig:compare}(A)).}
\rev{The vertical leap, 2.8 BL, is the highest reported for a fish-shaped robot or jumping
robot~\cite{chen2022tmech,chen2022ral,yu2023dolphin,dong2024iajm,xia2025rapidrecoil,gwon2023scale,chang2019jumping}
and is comparable to those of the archerfish~\cite{shih2017archerfish} and the
guppy~\cite{soares2013guppy} (Fig.~\ref{fig:compare}(B)).}
\rev{To our knowledge, the current platform is the most agile palm-scale robot to link free swimming with
leaping, offering a potential means to extend the operational range of underwater robots
beyond the submerged environment through aerial--underwater
transitions~\cite{zufferey2019jumpgliding,zufferey2026flapping}.}

\section{SYSTEM DESIGN OF THE ROBOT PLATFORM}
\label{sec:design}

\subsection{Mechanical Design}

\subsubsection{Hydrodynamics of the hull and rudder}
\label{sec:cfd}
\begin{figure}[t]
  \centering
  \includegraphics[width=0.92\columnwidth]{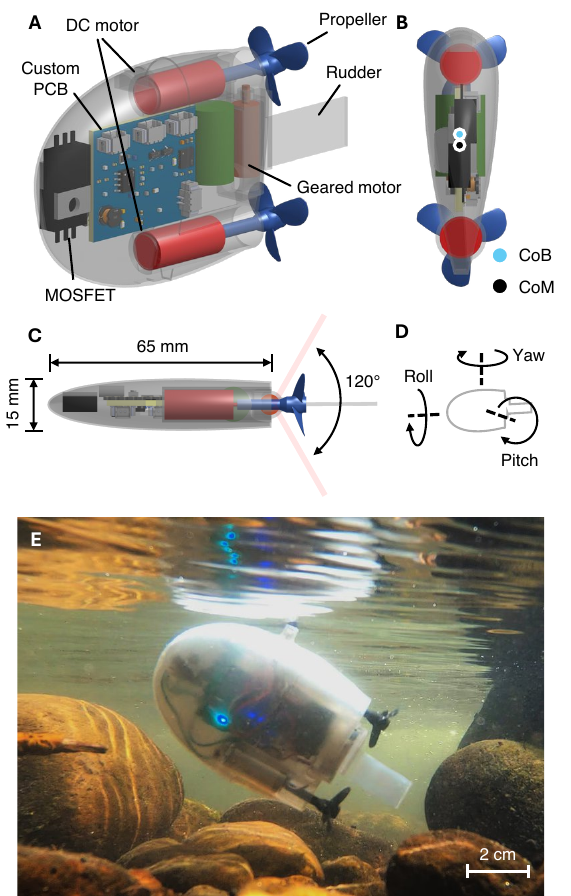}
  \caption{Robot design in (A)~oblique view with the sealed hull housing two
  coreless DC motors (one above the other in the center plane driving
  mirrored B-series propellers), the custom PCB, MOSFET drivers, and the
  geared rudder motor, (B)~front view with the center of buoyancy (CoB,
  cyan) and center of mass (CoM, black) markers, and (C)~top view (65~mm body length, 15~mm width)
  with the tail rudder sweeping $\pm$60 deg about the centerline. (D)~Body axes with the pitch, yaw, and roll rotations.
  (E)~The assembled prototype in a natural stream.}
  \label{fig:design}
\end{figure}

The current robot's body (Fig.~\ref{fig:design}) is a 3D-printed waterproof hull
of body length 65~mm \rev{(excluding the tail rudder and propellers)} and width 15~mm.
The current robot is neutrally buoyant at a total mass of 34~g, of
which the two motors account for 10.0~g, the electronics (PCB, MOSFET driver boards, and the bulk
capacitor) for 9.9~g, the printed body and covers for 5.6~g, the tail
assembly with its gearmotor for 1.8~g, and the propellers for 0.6~g.
The remaining 6~g is sealant and wiring.

The hull shape is a \rev{National Advisory Committee for Aeronautics (NACA)}
four-digit \rev{airfoil thickness form~\cite{abbott1959wing}} truncated at the
tail in all three views (Fig.~\ref{fig:design}) following the Kammback
principle, \rev{in which the tail ends abruptly just upstream of where the flow would}
\rev{separate~\cite{hucho1993road},} to keep both the drag and the body size small.
In each view the thickness curve is cut at the fraction $c$ of its chord and
stretched over the outline, and the (maximum width, length, $c$)
triples are (15~mm, 50~mm, 0.9), (50~mm, 65~mm, 0.5), and
(15~mm, 65~mm, 0.5) for the front, side, and top views, respectively.
A submerged body is passively stable in roll when its center of
buoyancy \rev{(CoB)} lies above its center of \rev{mass (CoM).} In the current robot the inverted airfoil
section of the front view places the displaced volume high, so the
\rev{CoB} lies 1.7~mm above and 0.7~mm forward of the \rev{CoM}
(Fig.~\ref{fig:design}(B)).

\begin{figure}[t]
  \centering
  \includegraphics[width=0.92\columnwidth]{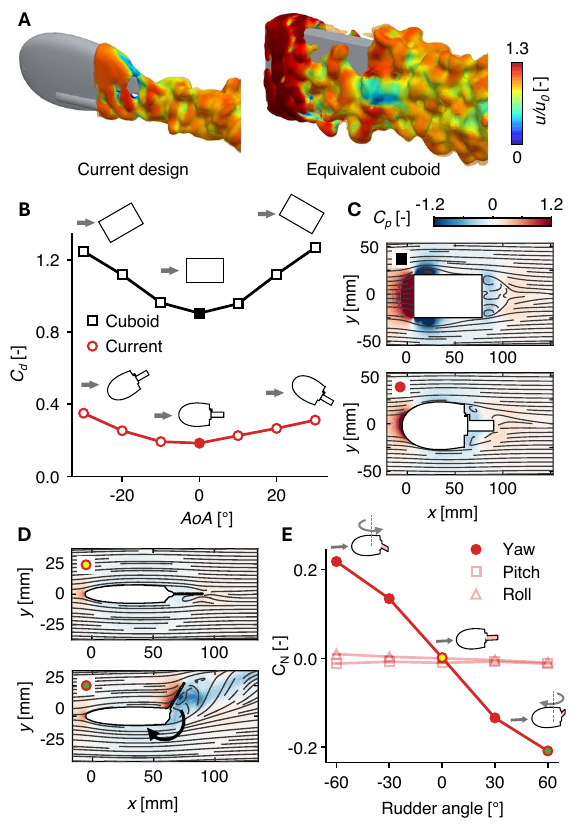}
  \caption{Hydrodynamics of the hull and rudder using computational fluid
  dynamics. (A)~Instantaneous wake structures behind the current design and behind the
  equivalent cuboid. (B)~Drag coefficient $C_d = F_x/(q A_f)$ against angle of attack from
  $-30$ deg to 30 deg for the current design and the cuboid, with $F_x$ the drag force,
  $q = \rho u_0^2/2$ the dynamic pressure at the free-stream speed $u_0 = 0.5$~m/s,
  and $A_f = 5.69$~cm$^2$ the frontal projected area shared by the current
  design and the cuboid. (C)~Contours of the pressure coefficient $C_p = (p - p_\infty)/q$ and
  streamlines around the cuboid and around the current design. (D)~Midplane flow
  with the rudder neutral and deflected. (E)~Yaw, pitch, and roll
  moment coefficients $C_N = M/(q A_f L)$ against rudder angle from
  $-60$ deg to 60 deg, with $M$ the moment about the \rev{CoM} and $L$ the
  overall length of the body with its rudder.}
  \label{fig:cfd}
\end{figure}

\rev{We compared} the hydrodynamic drag of the current design with that of the
equivalent cuboid (71~$\times$~13~$\times$~45~mm), a box with the same
frontal, side, and top projected areas as the current design
\rev{(Fig.~\ref{fig:cfd}(A)), using computational fluid dynamics (CFD) simulations}
\rev{that solve the incompressible unsteady RANS (URANS) equations with the}
\rev{$k$--$\omega$~SST model~\cite{menter1994sst} in OpenFOAM.}
At a free-stream speed of 0.5~m/s the current design sheds a narrow vortex train
confined to a cone behind the tail, whereas the cuboid sheds
relatively large vortices from its leading edges onward (Fig.~\ref{fig:cfd}(A),
iso-surfaces of the Q-criterion at $Q = 30$~s$^{-2}$ colored by the
local speed normalized by the free stream, $u/u_0$), which implies a
larger pressure drag and more flow energy lost for the cuboid. At zero angle of attack (AoA) the
current design has $C_d \approx 0.18$ against $\approx$ 0.9 for the
cuboid (Fig.~\ref{fig:cfd}(B)), roughly a five-fold reduction, as the cuboid’s flat front and sharp edges produce a large pressure difference between the front and separated wake (Fig.~\ref{fig:cfd}(C), top), while the current design keeps
the flow attached over most of its length and recovers pressure toward
the tail (Fig.~\ref{fig:cfd}(C), bottom). For angles of attack between $-30$ deg and 30 deg,
which occur during pitch maneuvers, the current design's $C_d$ stays at
$\approx$ 0.2--0.35, while the cuboid's stays at $\approx$ 0.9--1.3 (Fig.~\ref{fig:cfd}(B)).




\subsubsection{Vertically stacked thruster pair and tail rudder}
Two 8520 coreless DC motors (DaFuRui, 8.5~mm $\times$ 20~mm, 3.7~V) are
mounted vertically in the side-view center plane of the body (Fig.~\ref{fig:design}(A)--(C)). The thrust difference between the two
motors produces a
pitching torque with the
vertical offset as the lever arm, while the thrust sum sets the net
propulsive force. The two propellers are printed as mirror-image
twins (Section~\ref{sec:thrust}) so the pair can be mounted to contra-rotate
and the net propeller reaction torque about the roll axis cancels when
the two motors run at the same speed.

Yaw authority comes from a single flat tail rudder (25~mm long,
11~mm tall, 1~mm thick) mounted between the propeller pair.
The rudder is
driven by a 6~mm two-stage planetary gearmotor (Garosa, 26.45:1 reduction,
$\approx$ 2.6~mN$\cdot$m output torque)
through an H-bridge.
Rudder deflection creates a circulation that turns the body
(Fig.~\ref{fig:cfd}(D)), with a yaw-moment coefficient that grows
monotonically with angle over $\pm 60$ deg while the pitch and roll
moment coefficients stay negligible (Fig.~\ref{fig:cfd}(E)), so the
rudder steers the body in the horizontal plane without disturbing
the pitch PID control.
In straight free swimming, the flow passively realigns the rudder without position sensing. When the body yaws off target, a PID controller uses the gyro-integrated angle to drive the gearmotor with signed PWM and steer the body back toward the target.

\subsubsection{Propeller optimization}
\label{sec:thrust}

\begin{figure}[t]
  \centering
  \includegraphics[width=0.92\columnwidth]{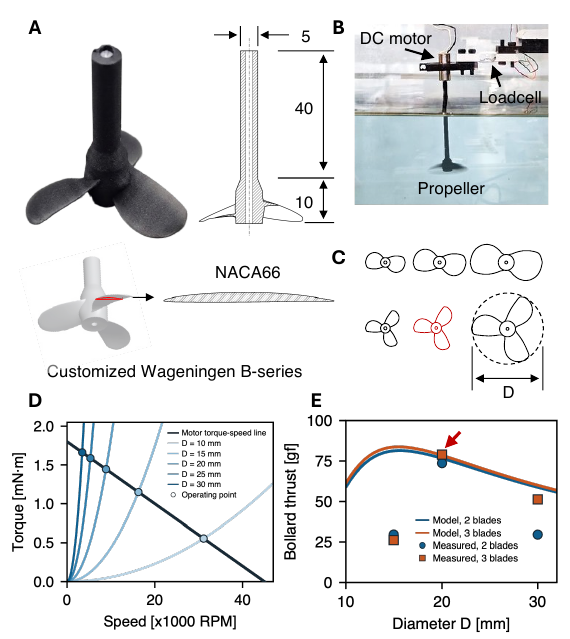}
  \caption{Propeller optimization. (A)~A printed
  three-blade customized Wageningen B-series propeller \rev{(modified NACA~66} blade
  sections, dimensions in mm). (B)~Bollard-thrust dynamometer, with the motor attached to a load
  cell and the propeller submerged in a water tank. (C)~Planforms
  of the printed variants, two- and three-blade at diameters $D$ from
  15 to 30~mm. The red outline is the selected three-blade 20~mm
  propeller, the one marked by the arrow in (E). (D)~The motor's torque--speed line against the
  B-series torque load curves $Q = K_Q \rho n^2 D^5$ for $D=10$--30~mm,
  with the operating points circled. (E)~Bollard thrust versus
  diameter $D$, predicted by the torque-matched model for two- and
  three-blade families (curves) and measured on the dynamometer (symbols).
  The arrow marks the selected three-blade 20~mm propeller (78.9~gf at
  2.95~A).}
  \label{fig:thrust}
\end{figure}

\rev{We optimized our propeller using the} Wageningen B-series model, a standard
marine propeller family~[\citen{kuiper1992wageningen}]--[\citen{carlton2019marine}],
because its geometry follows directly from a few parameters and its
thrust and torque are predictable. A B-series propeller is defined by
four parameters such as the blade count $Z$, the diameter $D$, the pitch ratio
$P/D$, and the expanded blade area ratio $A_E/A_0$, while the
others are fixed by the series, such as the chord, maximum thickness,
and skew distributions along the radius
(Fig.~\ref{fig:thrust}(A)).
The open-water thrust and torque coefficients,
$K_T = T/(\rho n^2 D^4)$ and $K_Q = Q/(\rho n^2 D^5)$ with $T$ the
thrust, $Q$ the shaft torque, and $\rho$ the water density, are
published as regression polynomials in the advance ratio $J$\rev{${}= V_a/(nD)$ (with $V_a$ the advance speed,
$n$ the rotation rate, and $D$ the propeller diameter)} and the
geometry.

At $J = 0$, corresponding to zero robot speed, $K_T/K_Q$ increases as the pitch ratio ($P/D$) decreases. A smaller blade area ratio ($A_E/A_0$) also reduces both blade torque and mass. Thus, for the torque-limited coreless motor, we fixed both at their lower bounds, $P/D = 0.6$ and $A_E/A_0 = 0.40$, leaving blade count ($Z$) and diameter ($D$) as the two remaining parameters.

For a given diameter, the rotation rate, $n$, follows from matching the torque required by the propeller to the torque supplied by the motor (Fig.~\ref{fig:thrust}(D)).
At rest the propeller requires
$Q = K_Q \rho n^2 D^5$, a parabola in the rotation rate $n$ with $K_Q$
from the B-series at $J = 0$, while the motor supplies torque along a
straight torque--speed line between the nominal 1.8~mN$\cdot$m
stall torque and 45~kRPM no-load speed of the 8520 class, and their
intersection, the operating point, fixes $n$.

With the pair of $n$ and $D$, the bollard thrust
$T = K_T \rho n^2 D^4$ follows, plotted in Fig.~\ref{fig:thrust}(E) for two and three
blades (solid curves in Fig.~\ref{fig:thrust}(E)) together with the dynamometer measurements (symbols in Fig.~\ref{fig:thrust}(E)).
In the bollard dynamometer (Fig.~\ref{fig:thrust}(B)), a single 8520 motor with the propeller submerged
is mounted on a 200~g load cell read by an amplifier (HX711), an inline sensor (INA260) records
the current, and a relay switches a regulated 5.0~V supply for five
pulses per propeller (5~s on, 20~s rest). \rev{We tested six printed propellers,} two- and three-blade
at $D = 15$, 20, and 30~mm (Fig.~\ref{fig:thrust}(C)), all at $P/D = 0.6$
and $A_E/A_0 = 0.40$.
As a result, the three-blade 20~mm propeller is the optimum, with a peak sustained thrust of 78.9~gf at 2.95~A, \rev{followed by} the two-blade 20~mm
propeller at 73.7~gf. At $D = 15$~mm, the measured thrust is substantially lower than the model predicts, at 29.6~gf for two blades and 26.1~gf for three blades. This discrepancy arises because the B-series blade thickness scales with diameter and falls below the printer’s 0.30~mm resolution limit at this small diameter. The printed blades are therefore thicker than specified by the series, so the B-series polynomials no longer accurately predict their performance.
At $D = 30$~mm the thrust drops to 29.5~gf (two blades) and
51.3~gf (three blades), because a larger $D$ steepens the torque parabola in
Fig.~\ref{fig:thrust}(D) and moves the operating point toward stall (i.e., $n \to 0$).
With both thrusters the static thrust-to-weight ratio is
$2 \times 78.9 / 34 \approx 4.7$\rev{. Against} the hull drag of Fig.~\ref{fig:cfd}(B) over
a one-body-length run-up from rest, that thrust would bring
the neutrally buoyant 34~g body to $v = 2.3$~m/s and a ballistic rise of
$v^2/(2g) \approx 0.27$~m, about 4 BL, which overestimates the measured height of 2.8 BL (Fig.~\ref{fig:compare}(B)) by neglecting added mass and the decrease in thrust at higher advance ratios ($J > 0$).

\subsection{Electronics and Firmware}
\label{sec:electronics}

\begin{figure}[t]
  \centering
  \includegraphics[width=\columnwidth]{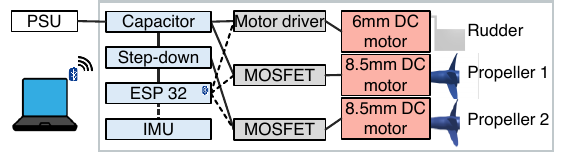}
  \caption{Electronics. Supply power passes through a bulk capacitor at
  the tether terminal and a step-down converter to the ESP32 and IMU.
  MOSFET channels and an H-bridge motor driver actuate the two 8.5~mm
  thrusters and the 6~mm geared rudder motor, respectively. A host PC connects
  over BLE for parameterization and telemetry only.}
  \label{fig:electronics}
\end{figure}

The custom board (32~$\times$~25~mm, 3~mm thick, $\approx$5~g,
Fig.~\ref{fig:electronics})
hosts an ESP32-C3 Super Mini module (AITRIP), an MPU-6050 IMU on I$^2$C (accelerometer
$\pm$2~g, gyro $\pm$250 deg/s, 42~Hz digital low-pass filter), IRLB8721
MOSFETs gating the thrusters, and a DRV8833 H-bridge for the rudder gearmotor.
Power enters at 12--20~V over a 1~m, 30~AWG twisted-pair tether and is stepped down
by an MP2307 synchronous buck to 5~V, from which the MCU module
regulates 3.3~V for logic and the IMU. A 1000~$\mu$F, 25~V
electrolytic at the tether terminal buffers the thrusters' current
transients.

\rev{We connect a host PC to the robot's board} over Bluetooth Low Energy (BLE) for
parameterization and $\approx$~70~Hz, 14-field telemetry (raw IMU,
per-motor PWM, sequence phase, estimated and target pitch, controller
output).
Once \rev{we start a run,}
the entire sequence executes on the current platform without
BLE communication while the board records an on-board log in RAM (raw
IMU, motor and rudder PWM, angles and targets at about 25~Hz, up to
about 120~s) and streams the log to the host PC over BLE when the run ends, so a
dropped link during the run \rev{does not lead to any data loss.}

\subsection{Closed-Loop Pitch Control}
\label{sec:pitch}


Pitch is estimated on-device by a time-based complementary
filter~\cite{mahony2008complementary} fusing the gyro rate with the
accelerometer gravity reference,
\begin{equation}
  \hat\theta_k = \alpha \left( \hat\theta_{k-1} + \omega\,\Delta t \right)
               + (1-\alpha)\,\theta_{\mathrm{acc}},
  \qquad \alpha = \frac{\tau}{\tau + \Delta t},
  \label{eq:filter}
\end{equation}
where $\hat\theta_k$ is the pitch estimate at step $k$, $\omega$ the
gyro pitch rate, $\Delta t$ the sample interval,
$\theta_{\mathrm{acc}}$ the accelerometer-derived pitch, and
$\tau = 2$~s the filter time constant. The accelerometer term is
rejected whenever the measured specific force deviates from 1~g by
more than 0.15~g (thrust and vibration corrupt the gravity
reference), leaving pure gyro integration for that step. The gyro bias
is updated at a low adaptation rate and only while the robot is idle, so a commanded rotation is not
absorbed into the bias estimate.

The controller is a PID on the pitch error $e$ between the estimate
$\hat\theta$ and the commanded pitch $\theta_{\mathrm{ref}}$,
\begin{align}
  e &= \hat\theta - \theta_{\mathrm{ref}}, \qquad
  u = K_p e + K_i \textstyle\int e\,dt + K_d\,\dot{\hat\theta},
  \label{eq:pid}\\
  \delta &= \mathrm{sat}_{[-\delta_{\max},\,\delta_{\max}]}(u), \qquad
  \begin{aligned}
    d_{\mathrm{top}} &= \mathrm{sat}_{[0,\,d_{\max}]}(d_{\mathrm{base}} + \delta),\\
    d_{\mathrm{bot}} &= \mathrm{sat}_{[0,\,d_{\max}]}(d_{\mathrm{base}} - \delta),
  \end{aligned}
  \label{eq:mixing}
\end{align}
where $\mathrm{sat}_{[a, b]}(x) = \min(\max(x, a), b)$ is the saturation
function, $u$ is the controller output, the trim $\delta$ (saturated at
$\pm\delta_{\max}$) tilts the thrust pair through the top- and
bottom-motor duties $d_{\mathrm{top}}$ and $d_{\mathrm{bot}}$, and the
base duty $d_{\mathrm{base}}$ sets the net thrust.

\begin{figure}[t]
  \centering
  \includegraphics[width=0.95\columnwidth]{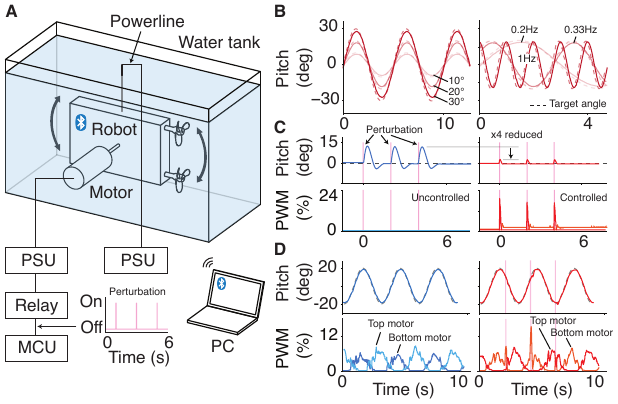}
  \caption{Validation of the closed-loop pitch control \rev{in a water tank.}
  (A)~The water tank setup in which the robot, here with an earlier
  rectangular hull, is mounted to a frame free to pitch, is
  powered from a \rev{laboratory power} supply through the tether, and reports
  telemetry over BLE to the host PC. A relay-driven motor on an independent
  supply serves as a free pivot when unpowered and delivers timed mechanical taps when the MCU (Arduino Uno) switches
  the relay (inset, three taps at 2~s spacing). (B)~Sinusoidal tracking,
  estimated pitch $\hat\theta$ against the commanded
  $\theta_{\mathrm{ref}}$ (dashed), across amplitudes 10--30 deg
  (left) and across frequencies 0.2--1~Hz (right). (C)~Impulsive
  disturbance rejection with the pitch commanded to 0 deg, pitch and
  per-motor PWM traces, uncontrolled (left) and controlled (right),
  with the tap instants marked by vertical pink lines. (D)~Taps delivered during sinusoidal
  tracking, pitch and per-motor PWM without (left) and with (right)
  perturbations.}
  \label{fig:pitch}
\end{figure}

\rev{We validated the PID control algorithm in a water tank,} first in sinusoidal tracking and
then in impulsive-disturbance \rev{rejection. The submerged robot was mounted}
\rev{at its CoM on the shaft of a pivot motor so that the robot could rotate only}
in pitch (Fig.~\ref{fig:pitch}(A)).
\rev{From the host PC, we} swept sinusoidal pitch commands over amplitudes of
10 deg, 20 deg, and 30 deg and frequencies of 0.2, 0.33, and 1~Hz
(Fig.~\ref{fig:pitch}(B)). \rev{For each run, the amplitude} $A$ and phase
$\varphi$ \rev{were extracted} by least-squares fitting
$A\sin(2\pi f t + \varphi)$ at the commanded frequency $f$ to the
$\approx$ 70~Hz telemetry, with the first period discarded, and every
metric \rev{was computed from the} raw data. The gains were optimized \rev{in this setup to} $K_p = 10$,
$K_i = 0$, \rev{and} $K_d = 0.5$ and held \rev{fixed} throughout, with a per-motor ceiling
$d_{\max} = 24\%$ here and in free swimming (Section~\ref{sec:sin})
and 100\% in the water exits (Section~\ref{sec:exit}).
The ratio of fitted to commanded amplitude stays at 0.88--0.96 across
every condition (Fig.~\ref{fig:pitch}(B)), and \rev{even with $K_i = 0$ the pitch returns}
\rev{to the command after small perturbations.}
The root-mean-square tracking error (RMSE) stays within 11\% of the
commanded amplitude over amplitudes of 10--30 deg and frequencies of
0.2--0.33~Hz, and grows to 30\% at 1~Hz, where the actuator begins to
saturate, due to phase lag rather than amplitude error (Fig.~\ref{fig:pitch}(B)).

\begin{figure}[t]
  \centering
  \includegraphics[width=0.95\columnwidth]{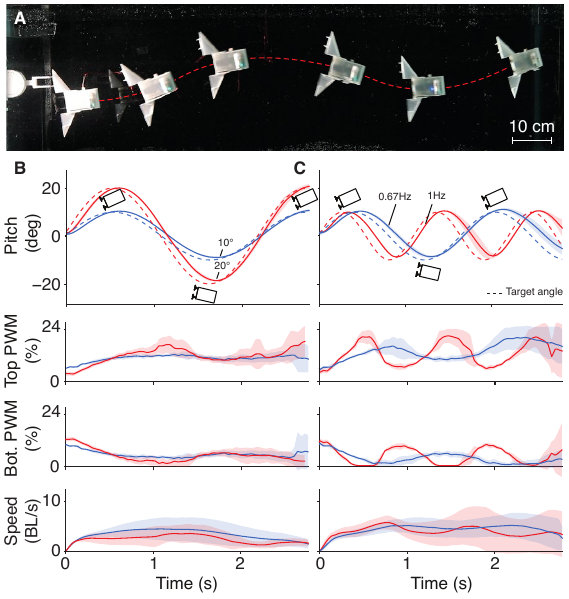}
  \caption{Free-swimming sinusoidal pitch tracking. (A)~Snapshot
  sequence of the $A = 10$ deg, 0.67~Hz swim.
  The dashed line traces the body center. (B)~Amplitude sweep, 10 deg (blue) and 20 deg (red) at
  0.5~Hz, and (C)~frequency sweep, 0.67~Hz (blue) and 1~Hz (red)
  at 10 deg. In both, the four rows are the estimated pitch
  $\hat\theta$ (solid) against the command $\theta_{\mathrm{ref}}$ (dashed), the
  top- and bottom-motor PWM duties, and the forward speed in BL/s, with shaded bands
  the standard deviation over $n = 5$ trials per case.}
  \label{fig:sin}
\end{figure}

\rev{We applied impulsive disturbances through the pivot motor}
\rev{(Fig.~\ref{fig:pitch}(A)), powered from an independent supply through} a relay
that an \rev{MCU switched,} so that the motor shaft \rev{tapped} the body directly in timed
10~ms pulses, three times at 2~s intervals.
With the controller off, the hull swings
$\pm12$ deg and rings down slowly (Fig.~\ref{fig:pitch}(C), left). With the PID commanded to $0$ deg, the excursions stay within $\pm3$ deg and recover within
about 1~s of each tap (Fig.~\ref{fig:pitch}(C), right), a
four-fold reduction of the swing. \rev{The tap instants detected} automatically from
gyro-rate spikes \rev{agree with} the programmed tap \rev{schedule.} In addition, taps delivered during
sinusoidal tracking push the instantaneous error to 6--8 deg but
are absorbed within about 0.3~s, leaving the overall RMSE
indistinguishable from undisturbed runs (Fig.~\ref{fig:pitch}(D)).
\section{UNDERWATER PITCH MANEUVERS}
\label{sec:sin}
\label{sec:pitchup}


\subsection{Sinusoidal Pitch Tracking}

\rev{In free swimming, we commanded the robot to follow} sinusoidal pitch at amplitudes of
10 deg and 20 deg and frequencies of 0.5, 0.67, and \rev{1~Hz.}
The robot follows the commanded sinusoid across all conditions
(Fig.~\ref{fig:sin}), with RMSE (mean $\pm$ standard deviation over
$n = 5$ trials per case) ranging from
$1.47$ $\pm$ 0.17 deg for the slowest command of 10 deg
amplitude at 0.5~Hz to $4.15$ $\pm$ 1.29 deg for the fastest command of
10 deg amplitude at 1~Hz.
The forward speed peaks at 4--6 BL/s at 24\% PWM
duty (Fig.~\ref{fig:sin}(B) and (C)). This sinusoidal steering demonstrates pitch control from the thruster
pair alone, with no dedicated dive planes or buoyancy controller. The runs used the earlier
rectangular hull of \rev{Fig.~\ref{fig:pitch}(A). Because these runs test} the
two-motor pitch loop rather than the \rev{hull, the change to the current} hull
is expected to \rev{alter} the drag and added mass \rev{but not the loop.}

\subsection{Pitch-Up}

\begin{figure}[t]
  \centering
  \includegraphics[width=0.95\columnwidth]{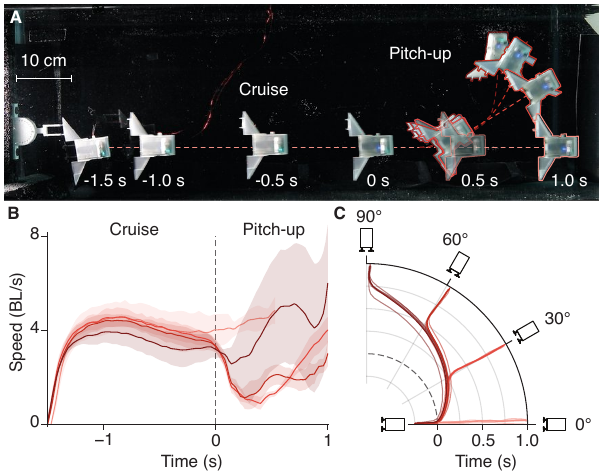}
  \caption{Pitch-up maneuver. (A)~Snapshot sequence of the straight
  approach ($t<0$) and the pitch-up starting at $t=0$, with overlaid
  trajectories fanning out to the four commanded target angles, 0, 30,
  60, and 90 deg. (B)~Forward speed through the cruise and the pitch-up for
  each target (mean over trials, shaded band the spread). (C)~Pitch $\hat\theta$ to each target in polar form, angle against
  time from the step (light traces individual trials, dark traces their
  mean), whose target labels also give the line colors in (A) and (B).}
  \label{fig:pitchup}
\end{figure}

The pitch-up maneuver rotates the swimming body from level to a
steep nose-up angle (Fig.~\ref{fig:pitchup}(A)).
While the robot swam straight, \rev{we stepped the target pitch to} 0, 30, 60,
or 90 deg \rev{nose-up for 1.5~s, and the on-board PID brought} the
robot to each commanded angle within 0.75~s (Fig.~\ref{fig:pitchup}(C)) and \rev{the robot retained} forward speed through the rotation
(Fig.~\ref{fig:pitchup}(B), $t > 0$).
The robot can therefore reach any
pitch angle from level to vertical in under a second, covering the
range needed for a water exit.

\section{WATER EXIT PERFORMANCE}
\label{sec:exit}

\subsection{Exit Sequence}

\begin{figure}[t]
  \centering
  \includegraphics[width=0.95\columnwidth]{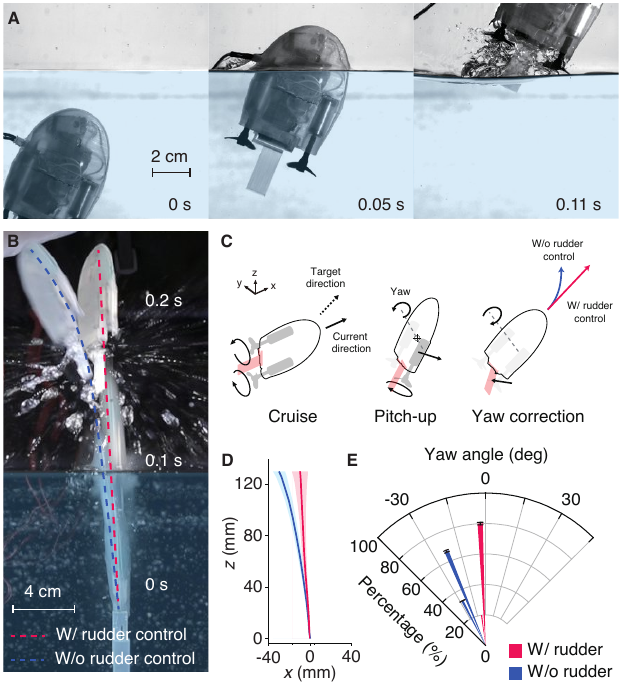}
  \caption{Water-exit maneuver. (A)~High-speed sequence of the breach in side
  view. (B)~Composite exit
  trajectories with (red dashed line) and without (blue dashed line) rudder control. (C)~Schematic of the
  exit sequence with the yaw torque of the thrust imbalance and the
  yaw correction by rudder control. (D)~Exit trajectories in the vertical
  plane, $z$ against $x$, with and without rudder control (mean line and shaded standard deviation, $n = 5$ trials per case). (E)~Polar histogram of the exit yaw angle over the same
  trials, with and without rudder control.}
  \label{fig:exit}
\end{figure}

\rev{We tested the} free-surface crossing and the rudder's yaw authority
during \rev{the crossing} from a start inclined at 65 deg
(Fig.~\ref{fig:exit}).
During the burst both
thrusters run at the commanded duty while the pitch PID control of
Section~\ref{sec:pitch} holds the pitch at the initial angle of 65 deg.
With the rudder yaw loop engaged, the
body breaks through the surface at the launch angle
(Fig.~\ref{fig:exit}(A)), leaving the water at 0.78~m/s, and the \rev{CoM}
rises 1.3 BL above the surface
(Fig.~\ref{fig:exit}(B), red dashed line).
With the yaw loop disengaged, the exit yaws out of the vertical launch plane
(see blue lines in Fig.~\ref{fig:exit}(B) and Fig.~\ref{fig:exit}(D)),
and across trials the exit yaw
angle reaches about 22 deg against about 3 deg with the loop engaged
(Fig.~\ref{fig:exit}(E)).

The pitch PID control holds the target angle by small, continual corrections
(Fig.~\ref{fig:exit}(C), left) that drive the two thrusters at unequal duties, leaving the reaction torques
of the contra-rotating propellers unbalanced. When the lower
thruster is the stronger, the uncanceled roll torque from the two motors can effectively be considered a lateral
force at the lower propeller, which lies behind the \rev{CoM,} and that force turns the body in yaw (Fig.~\ref{fig:exit}(C), middle). The rudder,
mounted between the propeller pair behind the \rev{CoM,} supplies the opposing
yaw moment with negligible pitch coupling (Fig.~\ref{fig:exit}(C), right),
so the exit stays in the launch plane even at exit speeds of
$\mathcal{O}(10)$~BL/s and in a turbulent stream with background flow
(Fig.~\ref{fig:field}).

\begin{figure*}[!t]
  \centering
  \includegraphics[width=0.95\textwidth]{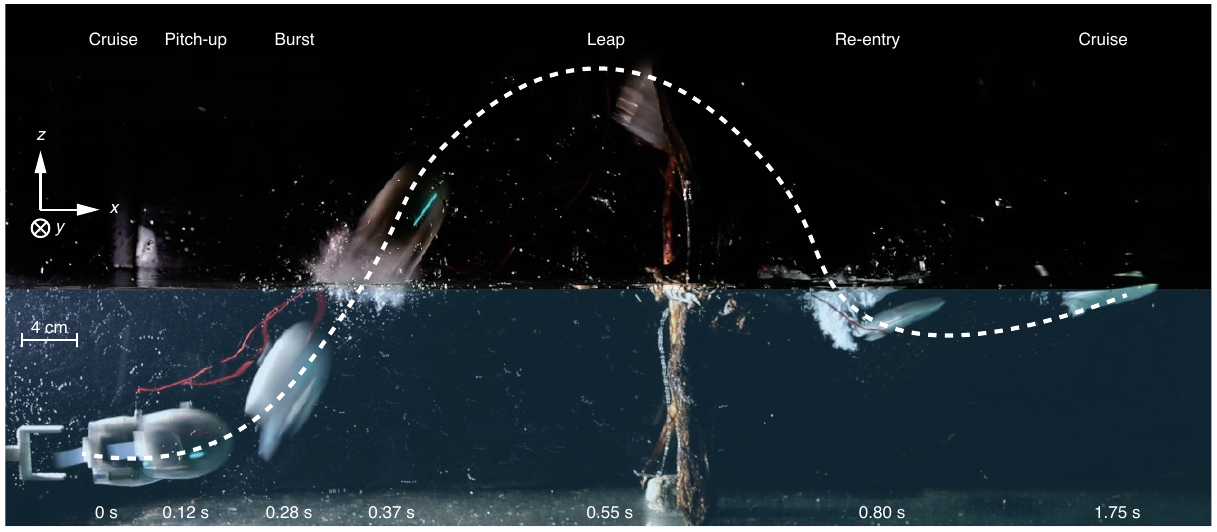}
  \caption{Complete leap over an obstacle in the tank, consisting of
  cruise, pitch-up, burst through the surface, ballistic flight over
  the obstacle, re-entry, and return to cruise, with the leap rising
  1.6~BL above the surface and spanning 3.7~BL.}
    \label{fig:leap}
\end{figure*}

\begin{figure}[t]
  \centering
  \includegraphics[width=0.95\columnwidth]{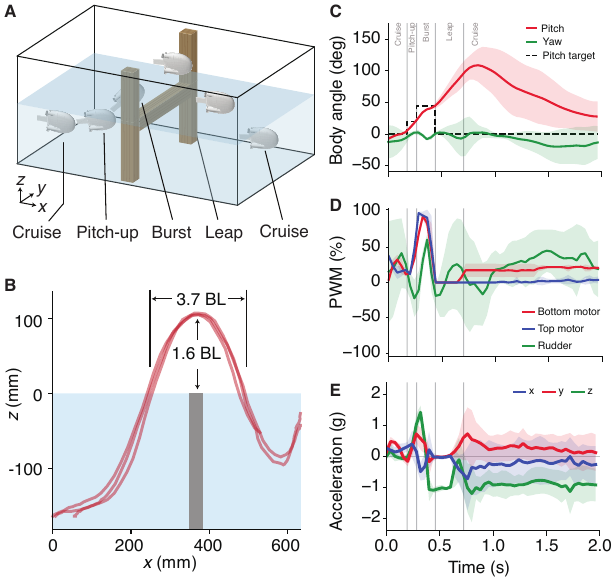}
  \caption{Leap kinematics and control signals in the tank. (A)~Schematic
  of the tank with the obstacle and the robot at the five phases:
  cruise, pitch-up, burst, leap with both thrusters
  off, and cruise again after re-entry. (B)~Body-center trajectories of
  three trials in the vertical plane, $z$ against $x$, the water shaded
  blue and the obstacle in gray, with the apex height and the leap length
  marked in BL. Time histories ($n = 5$ trials) of (C)~pitch
  and yaw against the stepped pitch target, (D)~thruster and rudder duty
  (both thrusters driven in the burst, motors off in the leap, low duty
  in cruise), and (E)~three-axis acceleration in the lab frame, with $z$
  the vertical axis along gravity. Vertical lines mark the phase boundaries, and
  shaded bands span trials.}
  \label{fig:leapdata}
\end{figure}

\subsection{Complete Leap}

Cruise, pitch-up, and burst, followed by a leap and a re-entry, comprise
a five-phase sequence run on-device
(Fig.~\ref{fig:leap} and the schematic of Fig.~\ref{fig:leapdata}(A))
whose phase boundaries are marked
by the vertical lines in Fig.~\ref{fig:leapdata}(C)--(E). In the leap
phase both thrusters switch off and the body flies ballistically with
the vertical acceleration in the lab frame at $-1$~g throughout
(Fig.~\ref{fig:leapdata}(E)), and cruise resumes after re-entry. In the tank the current robot cleared
an obstacle in all three trials, rising 1.6 BL above the surface and landing 3.7 BL horizontally from the takeoff point (Fig.~\ref{fig:leap} and
Fig.~\ref{fig:leapdata}(B)). With the pitch loop engaged throughout,
pitch follows the stepped nose-up target through the burst, continues past vertical in the air, and recovers
after re-entry, while yaw holds near zero
(Fig.~\ref{fig:leapdata}(C) and (D)).

\subsection{Field Demonstration}
\begin{figure*}[!t]
  \centering
  \includegraphics[width=0.95\textwidth]{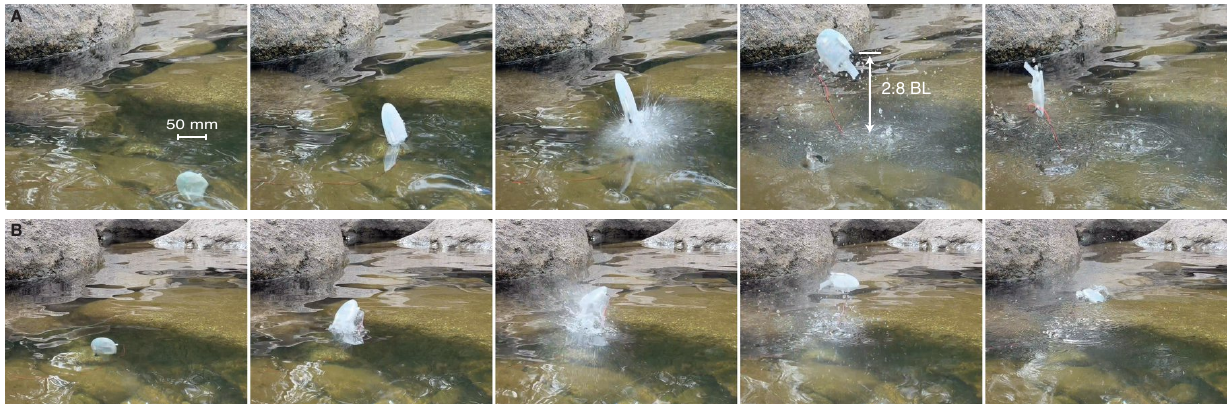}
  \caption{Water exits in
  \ifanon a natural stream\else \named{Boulder Creek, Colorado}\fi{}.
  (A)~An exit with rudder control, reaching 2.8~BL above the surface,
  from the surface start through the burst to the apex and the descent.
  (B)~An exit without rudder control, in which the body leaves the water
  near vertical, yaws away from the exit plane at the apex, and strikes the water
  surface far off along the yaw direction.}
  \label{fig:field}
\end{figure*}

\rev{We tested the} water exit outdoors in
\ifanon a natural stream\else \named{Boulder Creek, Colorado}\fi{}
(Fig.~\ref{fig:field}) to \rev{assess} whether the rudder keeps the exit
stable under turbulence and background flow. \rev{We launched the robot,}
powered through the tether from a portable battery, from a linear frame set underwater,
in a slow current with turbulence.
An exit with rudder control succeeded in the stream flow, reaching 2.8~BL above the surface
(Fig.~\ref{fig:field}(A)), which supports the robustness of the current design to external flow,
whereas an exit without rudder control left
the water near vertical, yawed away from the exit plane at the apex,
and struck the water surface far off along the yaw direction
(Fig.~\ref{fig:field}(B)), as in Fig.~\ref{fig:exit}(B).

\section{CONCLUSIONS}
\label{sec:conclusion}

We presented a palm-scale (65~mm, 34~g) swim-and-breach robot platform
whose two vertically stacked propellers provide propulsion and
differential-thrust pitch control, while a tail rudder adds yaw
control. In URANS simulation, the Kamm-cut NACA hull reduces the drag
five-fold against an equivalent cuboid, and rudder deflection yaws the
body with negligible pitch and roll moments. A three-blade 20~mm
propeller, selected among six candidates on the dynamometer, gives a static
thrust-to-weight ratio of 4.7, and the current robot swims at 13.9~BL/s
and turns at 209 deg/s, corresponding to the upper limits reported for underwater robots. The pitch loop tracked sinusoidal commands
with amplitude ratios of 0.88--0.96, held pitch within $\pm3$ deg
under impulsive disturbances, and reached nose-up angles up to
90 deg within 0.75~s. Rudder control stabilized the water exit,
reducing its yaw seven-fold, from about 22 deg to
3 deg. The current robot demonstrated the complete
cruise-leap-cruise sequence in the tank, leaping 1.6~BL high and
3.7~BL long. The rudder-controlled exit also reached 2.8~BL above the surface in a
turbulent natural stream, supporting the robustness of the current
design to external flow.

Future directions might include (i) untethered operation, through a joint
size--thrust--weight optimization with an onboard battery, a revised hull, and propeller optimization, (ii) a quantitative analysis of
underwater maneuvering performance in the horizontal plane (perpendicular to gravity) using rudder control, and (iii) repeated hop cycles with
controlled re-entry and attitude control in the air, where roll is unregulated, with systematic field trials in the habitats of the
fish that motivated the design.

The current robot platform shows the possibility of underwater robots crossing the weirs, rock sills, pipe lips, and dry gaps between pools that currently stop them in streams, flooded structures, and industrial systems, while providing a modular platform for environmental sensing, inspection, and monitoring.

\ifanon\else
\section*{\named{ACKNOWLEDGMENT}}
\named{S.~Bhamla acknowledges funding from Schmidt Sciences, LLC. D.~Choi acknowledges funding from the National Research
Foundation of Korea (\mbox{RS-2022-NR070924}, \mbox{RS-2023-00248034}).
The authors thank Paulami Sarkar (Bhamla Lab) for assistance with the experiments.}
\ifdefined\publishversion
\section*{\named{AUTHOR CONTRIBUTIONS}}
\named{Daehyun Choi: Conceptualization, Data curation, Formal analysis, Investigation, Methodology, Software, Validation,
Writing~-- Original Draft, Writing~-- Review \& Editing; Ian Bergerson: Conceptualization, Data curation, Formal analysis,
Investigation, Methodology, Software, Validation, Visualization, Writing~-- Original Draft; Hengjia Zhu: Data curation,
Formal analysis, Investigation, Validation, Visualization; Tianjun Lan: Data curation, Formal analysis, Investigation;
Saad Bhamla: Conceptualization, Funding acquisition, Resources, Supervision, Writing~-- Review \& Editing.}
\fi
\section*{\named{DATA AVAILABILITY}}
\named{The data that support the findings of this study are available from the corresponding authors upon reasonable
request.}
\fi


\end{document}